\documentclass{article} 
\usepackage{iclr2027_conference,times}

\usepackage{amsmath,amsfonts,bm}

\def\eqref#1{equation~\ref{#1}}

\def\1{\bm{1}}

\DeclareMathAlphabet{\mathsfit}{\encodingdefault}{\sfdefault}{m}{sl}
\SetMathAlphabet{\mathsfit}{bold}{\encodingdefault}{\sfdefault}{bx}{n}

\usepackage{hyperref}
\usepackage{url}

\title{QuacamFM: Quaternion-Constrained Flow Matching for Camera Pose Estimation}

\author{\small{Bao-Long Tran, 
        Cuong Le, 
        Tahereh Dehdarirad, 
         Fredrik Viksten, 
         Per-Erik Forss{\'e}n }\\
\small{Link{\"o}ping University, Sweden} \\
\small{\texttt{\{firstname.lastname\}@liu.se}}
}

\usepackage{xspace}
\newcommand{\ie}{\textit{i.e.}\xspace}
\usepackage[capitalize]{cleveref}
\usepackage{graphicx}
\usepackage{tabularx}
\usepackage{booktabs}
\usepackage{xfrac}
\usepackage[all,british]{foreign}
\usepackage[T1]{fontenc}

\newcommand{\btau}{\boldsymbol{\tau}}

\iclrfinalcopy 
\begin{document}

\maketitle

\begin{abstract}
Camera pose estimation from multi-view images remains a challenge in computer vision.
Traditional methods often address this problem using Structure-from-Motion (SfM) with bundle adjustment.
%
However, camera poses estimated from sparse views are inherently ambiguous due to insufficient geometric constraints.
Recent work leverages probabilistic models, such as diffusion models, to generate multiple camera pose hypotheses and therefore capture this uncertainty better. 
%
Most of these methods represent camera rotations using unit quaternions, but treat them as unconstrained 4D vectors during the generative processes, thereby ignoring the unit-norm constraint of quaternions.
Unconstrained quaternions create non-smooth and suboptimal generation trajectories.
%
To this end, we propose \emph{QuacamFM}, a quaternion-constrained flow matching framework for camera pose estimation that preserves unit quaternion representations throughout the entire flow trajectory.
We design the optimal transport of the quaternion flows using smooth spherical linear interpolation.
Experiments on CO3Dv2 demonstrate our method's advantage in camera pose accuracy over diffusion-based methods and classical SfM approaches.
We further show that our quaternion-constrained formulation outperforms the naive application of standard flow matching to 4D quaternion vectors on sparse-view camera pose estimation.
Finally, it is observed that QuacamFM generalizes well across datasets and in-the-wild examples.
%
\end{abstract}
\begin{figure*}[h]
    \centering
    \includegraphics[width=0.98\textwidth]{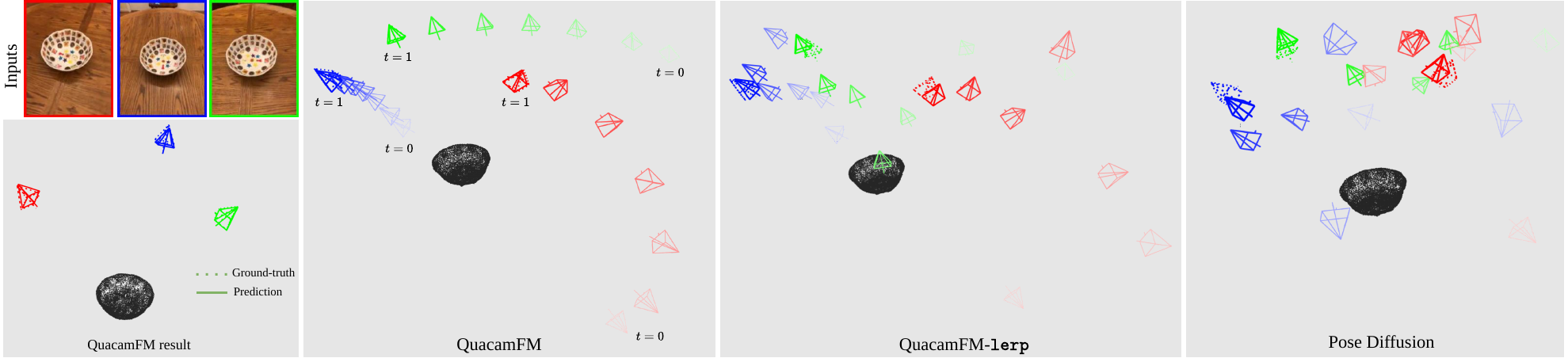}
    \caption{\textbf{Camera generation trajectory} of \textbf{QuacamFM} vs. QuacamFM-$\textup{\texttt{lerp}}$ and PoseDiffusion during inference, from starting step $t=0$ to ending step $t=1$. 
    Lighter shades indicate earlier flow steps, and darker shades indicate later steps.  
    QuacamFM produces smoothest trajectories, yielding better camera pose estimation accuracy.
    }
    \label{fig:teaser}
\end{figure*}
 
\section{Introduction}
\label{sec:intro}
Recovering camera poses, \ie translations and rotations from an unordered set of multi-view images is a fundamental problem in computer vision~\citep{rome}.
Accurate camera poses establish the essential geometric relationships in many downstream applications, including 3D reconstruction, visual localization, and robotic perception~\citep{orbslam2,vggsfm,duisterhof2025mastrsfm}. 
Traditionally, camera poses are estimated via Structure-from-Motion (SfM)~\citep{colmap_sfm}, which recovers camera geometry by exploiting geometric constraints on image correspondences and a bundle adjustment process in a global coordinate frame.
Recent advances in keypoint detection, feature matching, and correspondence estimation~\citep{superglue,roma,mast3r_eccv24} significantly strengthen SfM pipelines, especially under illumination changes. 

Despite these advances, SfM remains challenging in sparse-view settings, where limited image overlaps prevent the establishment of sufficiently reliable correspondences between image pairs.
Regression-based approaches overcome this by directly predicting camera poses or intermediate geometric representations from image observations without explicitly establishing pairwise correspondences, and have shown substantial improvements in sparse-view camera estimation~\citep{sparsepose,relpose,dust3r}.
However, deterministically regressing camera poses from sparse views is inherently ambiguous due to insufficient geometric constraints.
Probabilistic approaches have recently emerged as a promising alternative, particularly in challenging scenarios involving symmetric objects or image pairs with limited overlap~\citep{posediffusion,camera_as_rays,zhao2025diffusionsfm}, to better capture the uncertainty by generating multiple plausible camera pose hypotheses. 
Recently, PoseDiffusion~\citep{posediffusion} leverages the diffusion framework~\citep{ho2020denoising} to jointly model camera rotation and translation. 
Similar to~\cite{sparsepose,posenet,transformerpose}, PoseDiffusion uses 4D quaternions to represent camera rotations, but treating them as unconstrained 4D vectors and additively inject/remove noise during their generation.
The additive generation process breaks the unit-length constrain of quaternions, causing them to no longer represent pure 3D rotations, leading to decreases in camera rotation estimation accuracy.

In this paper, we propose \emph{Quaternion Camera Flow Matching} (QuacamFM), a novel probabilistic approach for sparse multi-view camera pose estimation, using an operation based on the Hamilton product to explicitly ensure the 3D-rotation validity of camera quaternion poses.
We learn the optimal-transport (OT) mapping from a source distribution, \eg the Gaussian distribution, to the target ground-truth camera quaternion pose distribution, and the velocity is modeled via a quaternion-valued differential equation (QDE).
When solving the QDE, unlike additions, the Hamilton product preserves unit-quaternions throughout the mapping trajectory by operating on the unit sphere manifold $\mathcal{S}^3$.
We define the desired optimal transport of a randomly sampled quaternion to its ground-truth target as the \emph{spherical linear interpolation} (\texttt{slerp}).
To further highlight our proposed method, we also consider a vanilla formulation, QuacamFM-\texttt{lerp}, that directly applies standard flow matching, with linear interpolation as the desired OT mapping, and the quaternions are normalized to unit norm at every integration step to yield valid rotations.
A demonstration of our proposed method showing smooth camera OT trajectory during inference can be seen in~\cref{fig:teaser}.
We train and evaluate our model on the CO3Dv2 dataset and show clear advantages of using \texttt{slerp} as OT trajectory over the vanilla \texttt{lerp} in camera rotation recovery.
We further assess its generalization on the held-out CO3Dv2 object categories and the scene-centric RealEstate10K dataset.
We also demonstrate the applicability of our method to in-the-wild self-captured data.
To summarize, our contributions are listed as follows:
\vspace{-4pt}
\begin{itemize}
    \item We propose QuacamFM, a novel quaternion-constrained flow-matching framework for sparse multi-view camera pose estimation that preserves valid camera rotations throughout the flow trajectory.
    \vspace{-3pt}
    \item We empirically show that enforcing the quaternion constraint in flow matching significantly improves camera rotation accuracy compared with naively treating quaternions as unconstrained 4D vectors in normalizing flows.
    \vspace{-3pt}
    \item Our method demonstrates significant improvements over previous methods, including classic SfM and diffusion-based approaches on CO3Dv2, while also generalizing well to unseen scenes from RealEstate10K and in-the-wild captures.
\end{itemize}
\section{Related work}
\label{sec:related_work}
\paragraph{Structure from Motion.} 
SfM is a long-standing problem in computer vision~\citep{hartley2003multiple,ozyecsil2017survey}, aiming to estimate camera parameters from image sequences through global~\citep{cui2015global,moulon2013global,wilson2014robust} or incremental~\citep{rome,colmap_sfm,snavely2006photo,wu2013towards} pipelines. 
Its principal components, including keypoint detection, feature matching, and reconstruction, have been substantially improved by learning-based techniques~\citep{dedode,roma,loma,hloc}.
Improvements in these individual components have boosted the overall performance of the SfM framework.
Feed-forward SfM methods have also recently emerged~\citep{vggsfm,vggt,light3r-sfm,zhang2025flare,duisterhof2025mastrsfm,dust3r} to directly learn 3D geometry rather than through iterative optimization.
DUSt3R~\citep{dust3r} and MASt3R~\citep{mast3r_eccv24} leverage large-scale training data to regress dense 2D-3D corresponding pointmaps, from which camera poses and various other geometric quantities can be recovered.
However, camera pose estimates in sparse-view unordered image settings still remain challenging due to limited pair-wise geometric constraints.
Regression-based approaches are proposed~\citep{posenet,sparsepose,brahmbhatt2018geometry} to learn the camera poses end-to-end.
Direct regression of camera rotation and translation has been explored by~\cite{posediffusion}, showing improved performance over the classical COLMAP pipeline when fewer than ten input views are available.
SparsePose~\citep{sparsepose} further improves upon naive pose regression by using the regressed camera poses as an initial estimate and subsequently refining them through a learnable iterative module that exploits 3D projective-consistent local features.
\paragraph{Probabilistic model for Camera pose estimation.} 
Probabilistic models have emerged as a promising approach for handling uncertainty in sparse-view settings.
RelPose~\citep{relpose} adopts an energy-based formulation to model distributions over relative camera rotations, enabling the estimation of a consistent set of camera rotations.
RelPose++~\citep{relpose++} builds upon this formulation with a more modern architecture and extends it to jointly estimate camera translations.
DiffusionSfM~\citep{zhao2025diffusionsfm} applies diffusion to optical rays and their endpoints to recover camera geometry.
Camera-as-Rays~\citep{camera_as_rays} formulates camera as distribution of optical rays in Plücker coordinates. 
PoseDiffusion~\citep{posediffusion} learns to denoise camera rotations and translations directly from random noise, complemented by geometric guidance during inference. 
However, PoseDiffusion formulates quaternions of the camera rotations as independent unconstrained 4D vectors.
We propose a novel method using flow-matching framework to explicitly preserve the quaternion constraint for sparse unordered multi-view camera pose estimation.
\begin{figure}
    \centering
    \includegraphics[width=0.99\linewidth]{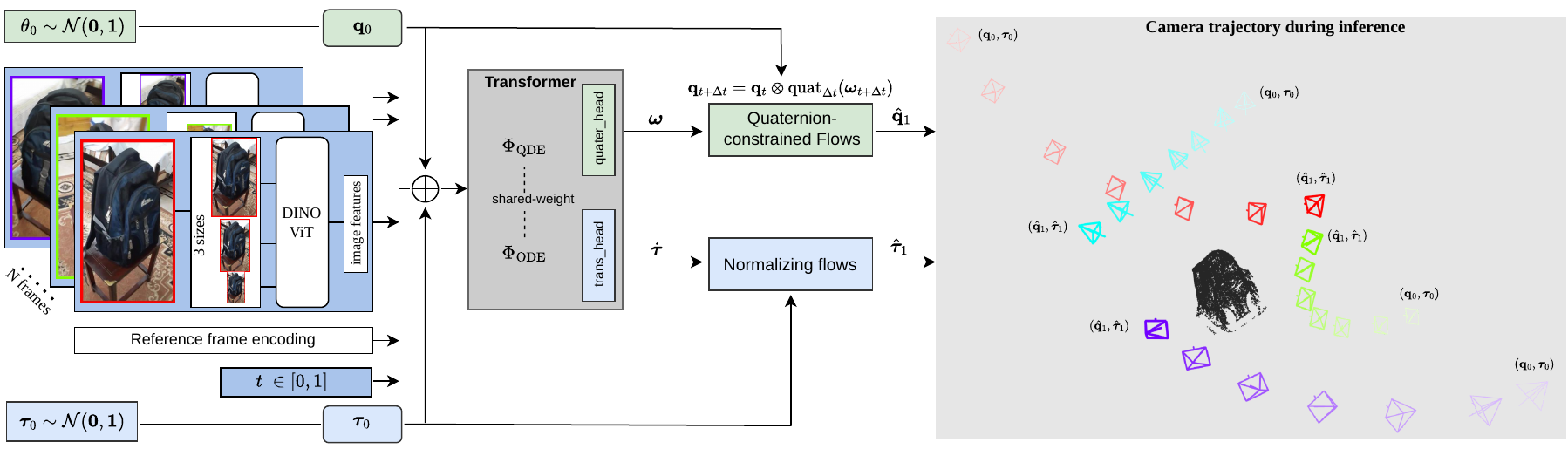}
    \caption{\textbf{QuacamFM overview}. We present a flow-matching-based method to predict camera poses given an image sequence. 
    The camera rotation is estimated via quaternion-constrained flows, where the optimal transport path is constructed using \texttt{slerp}, preserving the unit-quaternion constraint.
    The quaternion trajectory is integrated using the quaternion differential equation, while the translation uses a standard normalizing flow. 
    The rightmost picture shows the camera trajectory during inference, lighter shades indicate earlier flow steps, and darker shades indicate later steps.}   
    \label{fig:overview}
\end{figure}
\section{Method}
\label{sec:method}
From an image sequence $I=(I^i)_{i=1}^{N}$ of a scene, where $N\in\mathbb{N}$ is the number of input images and each image $I^i\in\mathbb{R}^{3\times H\times W}$, we estimate the camera extrinsics, \ie camera translations and rotations.
Each corresponding camera pose is represented as $(\mathbf{q}^i,\btau^i)$, where a unit quaternion $\mathbf{q}^i\in\mathbb{R}^4$ denotes the camera rotation and $\btau^i\in\mathbb{R}^3$ denotes the camera translation.
The camera extrinsic parameterization in the canonical coordinate system is detailed in~\cref{subsec:method-detail}.

\cref{fig:overview} presents the overview of our quaternion-constrained flow matching framework.
For camera rotation, we learn the optimal transport (OT) angular velocity, \ie quaternion flows, to map randomly sampled camera rotations to their corresponding ground-truths, under the unit-sphere constraint of quaternions.
For camera translation, we learn the OT velocity in Euclidean space, \ie normalizing flows, for the translational mapping.
Details are described in the following sections.

\subsection{Quaternion Differential Equation}
\label{subsec:qde}
A quaternion $\mathbf{q}\in\mathbb{R}^4$ is described by a 4D vector $(\mathrm{w}, v_1, v_2, v_3)$, expressed as:
\[ \mathbf{q} = \mathrm{w} + v_1i + v_2 j + v_3 k  = (\mathrm{w}, \mathbf{v})\, , \]
where $i,j,k$ are imaginary units satisfying $i^2=j^2=k^2=-1$. 
To correctly represent a 3D rotation, this 4D vector must be a unit quaternion, \ie normalized to unit length, $\lVert \mathbf{q} \rVert = 1$.
We refer to all unit quaternions as simply quaternions for convenience in this paper.
The Hamilton product $\otimes$ combines two quaternions, $\mathbf{q}_1=(\mathrm{w}_1,\mathbf{v}_1)$ and $\mathbf{q}_2=(\mathrm{w}_2,\mathbf{v}_2)$, into a new quaternion as \mbox{$\mathbf{q}_1 \otimes \mathbf{q}_2 = \left( \mathrm{w}_1\mathrm{w}_2 - \mathbf{v}_1^{\top} \mathbf{v}_2, \mathrm{w}_2\mathbf{v}_1 + \mathrm{w}_1\mathbf{v}_2 + \mathbf{v}_1 \times \mathbf{v}_2 \right)$}.
The Hamilton product is non-commutative, \ie $\mathbf{q}_1\otimes\mathbf{q}_2\neq\mathbf{q}_2\otimes\mathbf{q}_1$.
Following prior work~\citep{le2026quamo,golabek2022aircraft,kuipers1999quaternions}, the quaternion flow is described by a quaternion-value ordinary differential equation (QDE) as:
\begin{equation}
    \label{eq:qde}
    \dot{\mathbf{q}} = \frac{1}{2}
    \begin{bmatrix}
        -[\boldsymbol{\omega}]_\times & \boldsymbol{\omega} \\ -\boldsymbol{\omega}^\top & 0 \\
    \end{bmatrix}
    \mathbf{q}, \quad
    [\boldsymbol{\omega]}_\times =
    \begin{bmatrix}
        0 & -\omega_3 & \omega_2 \\
        \omega_3 & 0 & -\omega_1 \\
        -\omega_2 & \omega_1 & 0
    \end{bmatrix},
\end{equation}
where \(\boldsymbol{\omega}\in\mathbb{R}^3\) is the angular velocity vector, whose direction and magnitude represent the rotation axis and angular speed, respectively.

\paragraph{Solution to the QDE.} 
In conventional flow matching, the state is evolved by integrating the predicted velocity field using additive ordinary differential equation (ODE) updates.
However, for quaternion flows, the quaternion representation of a camera rotation $\mathbf{q}_t$ is constrained to the manifold $\mathcal{S}^3$, and a vanilla additive ODE update does not preserve the unit-norm constraint.
Therefore, to solve the QDE numerically, the Hamilton product is required to integrate the predicted velocity field while ensuring the quaternion solution remains on $\mathcal{S}^3$.
We implement multiplicative quaternion integration via the Hamilton product as:
\begin{equation}
    \label{eq:solution_qde}
    \mathbf{q}_{t+\Delta t}
    = \mathbf{q}_{t} \otimes \textup{quat}_{\Delta t}(\boldsymbol{\omega}_{t+\Delta t})
    = \mathbf{q}_{t} \otimes
    \left(
        \cos\left(\frac{\theta}{2}\right),
        ~\sin\left(\frac{\theta}{2}\right)
        \frac{\boldsymbol{\omega}_t}{\lVert \boldsymbol{\omega}_t\rVert}
    \right)~,
\end{equation}
where $\theta=\lVert \boldsymbol{\omega}_t\rVert \Delta t$ is the scalar magnitude of the rotation scaled by a time step $\Delta t$, $\boldsymbol{\omega}_{t}$ is normalized to a unit vector, and $\textup{quat}_{\Delta t}(\cdot)$ is the quaternionized angular velocity.
This integration scheme preserves the unit-norm constraint of quaternions on $\mathcal{S}^3$, resulting in smooth and continuous generation trajectories for camera rotations.
We use this integration in our QuacamFM, and~\cref{sec:experiment} shows superior results in camera rotation accuracy compared to naively using normalizing flows, see~\cref{fig:slerp_lerp}.

\subsection{Flow Matching for Camera}
\label{subsec:fm_camera}
From the input images $I$, QuacamFM learns to jointly estimate the OT rotation and translation velocities to recover
the ground-truth camera rotations $\mathbf{q}_1$ and camera translations $\boldsymbol{\tau}_1$.
We use the extracted visual features $\psi(I)$ from DINO-ViT~\citep{dino} as the generative flow condition, and the details of this extraction, including how we handle the coordinate ambiguity, are provided in~\cref{subsec:method-detail}.
The estimation of camera rotation and translation are described in the following sections, see also~\cref{fig:overview} for an overview.

\paragraph{Quaternion Flow Matching.} 
Given the visual features of image inputs $\psi(I)$, we seek to learn the optimal transport from random to the target camera quaternion rotation.
Consider the mapping over the time interval $t\in[0,1]$ from an initial random quaternion sample $\mathbf{q}_0$ to the ground-truth target quaternion $\mathbf{q}_1$, the diffeomorphic flow from $\mathbf{q}_0$ to $\mathbf{q}_1$ is defined via the QDE in~\cref{eq:qde}, and the angular velocity is estimated as:
\begin{equation}
    \label{eq:para_qde}
    \boldsymbol{\omega}_t = \Phi_{\textup{QDE}}(\mathbf{q}_t,t,\psi(I))~,
\end{equation}
where $\Phi_{\textup{QDE}}$ is the parameterized model, here chosen as the Transformer architecture~\citep{transformer}, and $t\in [0,1]$ is the timestamp.
Common flow matching pipelines~\citep{flowmatching,tong2023improving,zhang2025towards} often adopt a linear interpolation between source and target samples as the OT conditional probability path, due to its simplicity and stable training behavior.   
As explained in~\cref{subsec:qde}, quaternions must stay on the unit sphere $\mathcal{S}^3$ to correctly represent rotations, we identify the OT path as a \emph{spherical linear interpolation} (\texttt{slerp}) when learning camera rotations. 
The interpolated value $\mathbf{q}_t$, using \texttt{slerp} between $\mathbf{q}_0$ and $\mathbf{q}_1$, at time $t$ is defined as:
\begin{equation}
    \label{eq:slerp}
    \mathbf{q}_t = \textup{\texttt{slerp}}(\mathbf{q}_0,\mathbf{q}_1,t) = \mathbf{q}_0 \otimes (\mathbf{q}_0^{-1} \otimes {\mathbf{q}}_1)^{t}~.
\end{equation}
To instantiate a quaternion $\mathbf{q}_0$ for a valid 3D rotation, we sample a random axis-angle rotation $\boldsymbol{\theta}_0\in \mathbb{R}^3$ from the Gaussian distribution, $\boldsymbol{\theta}_0\sim\mathcal{N}(0,\mathbf{I}_3)$, and then convert it to quaternion $\mathbf{q}_0$.
Following that, the OT quaternion derivative $\dot{\mathbf{q}}^*$ is calculated as:
\begin{equation}
    \label{eq:quat_derivative}
    \dot{\mathbf{q}}^* = \log(\mathbf{q}_0^{-1} \otimes \mathbf{q}_1)~,
\end{equation}
where $\boldsymbol{\omega}^*$ is the corresponding angular velocity, derived by taking only the vector part of $\dot{\mathbf{q}}^*$~\citep{sola2017quaternion,dunn2011primer}. 
Note that \textup{\texttt{slerp}} is a linear interpolation with a constant derivative, thus $\boldsymbol{\omega}^*$ is independent of time.
We learn to predict the angular velocity $\boldsymbol{\omega}_t=\Phi_{\textup{QDE}}(\mathbf{q}_t,t, \psi(I))$ via a regression objective:
\begin{equation}
    \label{eq:loss-qde}
    \mathcal{L_\omega} = \lVert \boldsymbol{\omega}_t - \boldsymbol{\omega}^* \rVert_1 ~, \, \forall t \in [0, 1]~.
\end{equation}

\paragraph{Vanilla Quaternion Transport.} To highlight the importance of quaternion OT flow matching, we also provide a variant (QuacamFM-\textup{\texttt{lerp}}) using the vanilla linear transport \texttt{lerp}, which treats the camera quaternion rotation as an un-constrained 4D vector. 
We keep other components unchanged, but the interpolated quaternion and the ground-truth quaternion derivative in~\cref{eq:slerp} and~\cref{eq:quat_derivative} are replaced with:
\begin{equation}
    \label{eq:lerp}
    \begin{split}
    & \mathbf{q}_t = \textup{\texttt{lerp}}(\mathbf{q}_0,\mathbf{q}_1,t) = (1-t)\mathbf{q}_0 + t~\mathbf{q}_1~, \\
    & \Rightarrow \dot{\mathbf{q}}^*_{~lerp} = \mathbf{q}_1 - \mathbf{q}_0~.
    \end{split}
\end{equation}
The angular velocity $\boldsymbol{\omega}^*$ calculation and learning objective remain the same as when using \textup{\texttt{slerp}}.

\paragraph{Translation Flow Matching.} 
Following the common flow matching formulation for unconstrained variables, we model the translation using a linear interpolation (\textup{\texttt{lerp}}) conditional path and an additive ODE solver.
The initial sample $\boldsymbol{\tau}_0 \in \mathbb{R}^3$ is sampled from the Gaussian distribution, $\boldsymbol{\tau}_0\sim\mathcal{N}(0,\mathbf{I}_3)$. 
The transport trajectory and the OT translation velocity are then given by:
\begin{equation}
    \label{eq:lerp_trans}
    \begin{split}
    & \boldsymbol{\tau}_t = \textup{\texttt{lerp}}(\boldsymbol{\tau}_0, \boldsymbol{\tau}_1,t) = (1-t)~\boldsymbol{\tau}_0 + t ~\boldsymbol{\tau}_1~, \\
    & \Rightarrow \dot{\boldsymbol{\tau}}^* = \boldsymbol{\tau}_1 - \boldsymbol{\tau}_0~.
    \end{split}
\end{equation}
The transport velocity estimator of the translation vector is a regression model \mbox{$\dot{\boldsymbol{\tau}}_t=\Phi_\textup{ODE}(\boldsymbol{\tau}_t, t, \psi(I))$}, supervised by the learning objective:
\begin{equation}
    \label{eq:loss-ode}
    \mathcal{L}_{\dot{\boldsymbol{\tau}}} = \lVert \dot{\boldsymbol{\tau}}_t - \dot{\boldsymbol{\tau}}^* \rVert _1~, \, \forall t \in [0, 1]~.
\end{equation}
\subsection{Solving the camera extrinsic.}
Once the transport velocities $\boldsymbol{\omega}$ and $\dot{\boldsymbol{\tau}}$ are estimated, the camera rotation and translation can be solved numerically.
In particular, the quaternion vector and translation vector are integrated separately, corresponding to the QDE in~\cref{eq:qde} and an ODE.
Following Euler integration scheme, we solve the camera rotation via the numerical integration with predicted $\boldsymbol{\omega}$, expressed as:
%
\begin{equation}
    \label{eq:rk2_qde}
        \mathbf{q}_{t+\Delta t} = \mathbf{q}_t \otimes \textup{quat}_{\Delta t}(\boldsymbol{\omega})~,
\end{equation}
where $\otimes$ denotes the Hamilton product, $\psi(I)$ is the input image feature, and $\textup{quat}_{\Delta t}(\boldsymbol{\omega})$ is the quaternionized angular velocity following~\cref{eq:solution_qde}.
The camera translation is obtained by integrating the ODE using the Euler solver:
\begin{equation}
    \label{eq:rk2_ode}
    \boldsymbol{\tau}_{t+\Delta t} = \boldsymbol{\tau}_t + \dot{\boldsymbol{\tau}}\Delta t.
\end{equation}
At the end of these integration procedures, we acquire the estimated quaternion representation $\hat{\mathbf{q}}_1$ of the camera rotation and the estimated camera translation $\hat{\boldsymbol{\tau}}_1$.
Note that, in the variant \mbox{QuacamFM-\textup{\texttt{lerp}}}, which treats quaternion as an un-constrained 4D vector and uses vanilla \textup{\texttt{lerp}} as the optimal transport, the integration is updated identically as~\cref{eq:rk2_ode} but replacing $\boldsymbol{\tau}$ with $\mathbf{q}$.

The total learning objective is the aggregation of all the transport velocity loss in~\cref{eq:loss-qde}, ~\cref{eq:loss-ode}, and the reconstructed camera extrinsic loss over camera poses.
In which, the camera extrinsic losses, $\mathcal{L}_\mathbf{q}$ and $\mathcal{L}_{\boldsymbol{\tau}}$, are computed against the ground-truth poses, supervising the estimated rotation and translation, respectively.
With the hyperparameters $\lambda_\omega$, $\lambda_{\dot{\boldsymbol{\tau}}}$, $\lambda_{\mathbf{q}}$, $\lambda_{\boldsymbol{\tau}}$ balance the contributions of the loss terms, the total loss is computed as:
\begin{equation}
    \label{eq:loss}
    \mathcal{L} = \lambda_\omega \mathcal{L}_{\omega} + \lambda_{\dot{\boldsymbol{\tau}}} \mathcal{L}_{\dot{\boldsymbol{\tau}}} + \lambda_{\mathbf{q}} \mathcal{L}_{\mathbf{q}} + \lambda_{\boldsymbol{\tau}} \mathcal{L}_{\boldsymbol{\tau}}~.
\end{equation}

\subsection{Implementation details}
\label{subsec:method-detail}
\paragraph{Coordinate Ambiguity.} 
Inspired by the work of~\cite{posediffusion}, to avoid overfitting to a certain scene coordinate, we randomly select $N$ input images, and use one of them as the reference.
This means the reference camera is transformed to the origin, resulting in an identity rotation.
We transform the inputs to canonical form with respect to the reference-camera transformation.
This preserves the relative poses of the input cameras with respect to the randomly selected reference frame.
Since camera translation is only determined up to an arbitrary scale, we normalize the translations by dividing them by the median of the norms of the transformed translation vectors, same as the variant Quacam-\textup{lerp}.
To further prevent over-fitting to a certain number of input frames, we train with dynamic input frames, \ie randomly chose between 3-30 frames at each training step.

\paragraph{Experimental Setting.} 
We use the pre-trained DINO ViT~\citep{dino} to extract input image features $\psi(I)$. 
Following the setup of~\cite{posediffusion}, we crop the input images around the object center and resize them to $224\times224$.
We then extract visual features at multiple scales of $1, \sfrac{1}{2}$, and $\sfrac{1}{3}$ and average the resulting features.
The weights of DINO are enabled to be optimized during training.
To allow the model to identify the selected reference frame, we concatenate its extracted visual feature with a one-hot indicator set to $1$ along the last channel.
We adopt the Transformer architecture from~\cite{transformer} for the shared-weight transport velocity estimator of $\Phi_{\textup{ODE}}$ and $\Phi_{\textup{QDE}}$. 
The estimator has $8$ encoder layers, with $8$ attention heads, and the hidden dimension of the feedforward network is $1024$.
We use the multi-layer perceptron (MLP) for the quaternion head and translation head in~\cref{fig:overview}.
The hyper-parameters are set to $\lambda_\omega=1.0$, $\lambda_{\dot{\boldsymbol{\tau}}}=1.0$, $\lambda_{\mathbf{q}}=0.1$, $\lambda_{\boldsymbol{\tau}}=2.0$.
To accelerate the training process, we repeat input batches, allowing the model to see a batch at multiple random timesteps at once.
\begin{figure*}[t]
    \centering
    \includegraphics[width=0.99\textwidth]{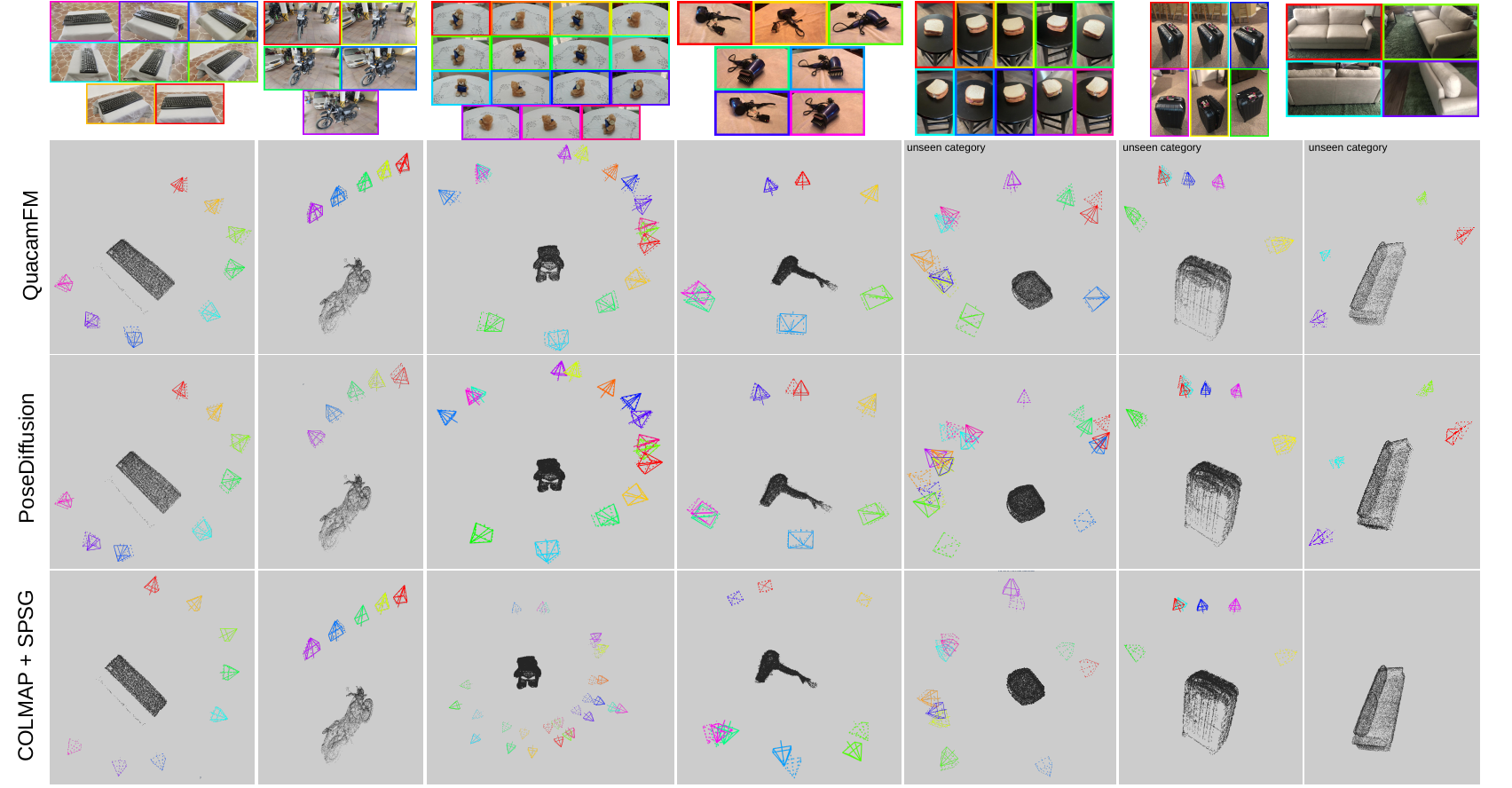}
    \caption{\textbf{Qualitative results on CO3Dv2}. 
    Camera pose estimations from input images (top row).
    Our QuacamFM and baseline results are shown in below rows.
    Camera frustum colors correspond to input-image borders.
    Solid frustums are camera estimations and dash frustums are camera ground-truths.
    The first four examples are from seen categories, and last three from held out (unseen) categories.
    Missing cameras indicate non-convergence.}
    \label{fig:qualitative}
\end{figure*}
\section{Experiments}
\label{sec:experiment}
\paragraph{Datasets.}
We use two datasets with different characteristics. 1) The object-centric CO3Dv2 \citep{co3d}, containing about 37k turntable-style videos from 51 MS-COCO categories~\cite{coco}, with camera poses estimated by COLMAP~\citep{colmap_sfm} from 200 frames. 
Following~\cite{relpose,posediffusion}, we train on 41 categories only, and hold-out the remaining 10 categories for evaluating generalization to unseen object categories. 
2) The scene-centric RealEstate10K~\citep{re10k}, with around 80k YouTube videos of indoor and outdoor scenes; its camera trajectories are estimated by ORB-SLAM2~\citep{orb-slam} and refined with bundle adjustment. Following~\cite{wiles2020synsin,posediffusion}, we evaluate on randomly sampled 1.8k videos from the original 7k-video test set, without further training. 
%
%
%
\begin{figure*}[t]
    \centering
    \includegraphics[width=0.99\textwidth]{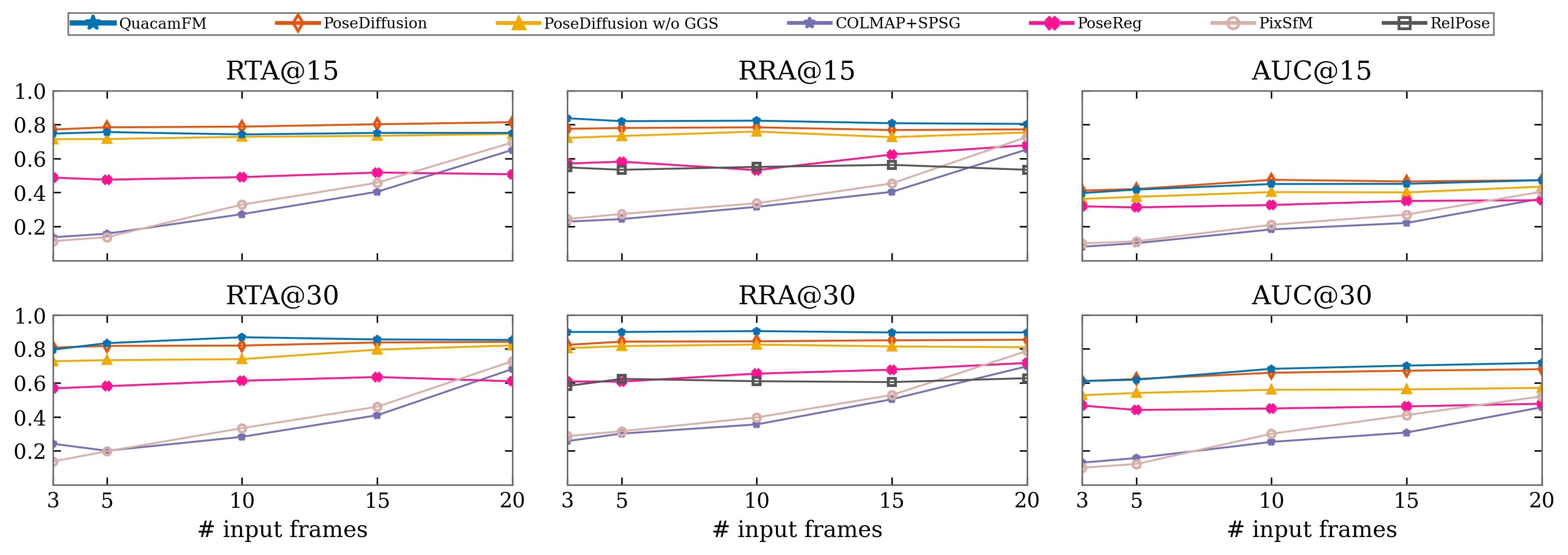}
    \caption{\textbf{Camera pose accuracy on CO3Dv2}. Quantitative metrics ($y$-axis, higher-better) at different thresholds and numbers of input images ($x$-axis). RelPose does not predict camera translation and hence is omitted in the respective metrics.}
    \label{fig:quantitative_co3d}
\end{figure*}
\begin{table*}[t]
    \footnotesize
    \small
    \setlength\tabcolsep{1.5pt}
    \caption{\textbf{Results on CO3Dv2}. Quantitative evaluation of camera pose estimation with 10-frame input.
    \textbf{Bold} denotes the best and \underline{underline} the second best.
    The inference time is measured in seconds $(\mathrm{s})$.
    \textsuperscript{\dag}Methods that require correspondences or post optimization.
    }
    \label{tab:result_co3d}
    \centering
    \begin{tabularx}{0.99\linewidth}{l@{\extracolsep{\fill}}ccccccccccccc}
        \toprule
        {Method} & {RTA@30$\uparrow$} & {RRA@30$\uparrow$} & {AUC@30$\uparrow$} & {RTA@15$\uparrow$} & {RRA@15$\uparrow$} & {AUC@15$\uparrow$} & $\mathrm{s}$\\
        \midrule
        PixSfM &  40.4 & 38.6 & 30.1 & 32.9 & 33.7 & 28.3 & 20.2 \\
        COLMAP + SPSG & 28.2 & 44.4 & 25.1 &  27.3 & 31.6 & 21.1 & 3.26\\
        PoseReg & 61.3 & 65.5 & 45.5 & 49.1 & 53.2 & 35.7 &  0.03\\
        RelPose\textsuperscript{\dag} & - & 72.7 & - & - & 57.2 & - &  29.5 \\  
        RelPose++\textsuperscript{\dag} & 84.3 & \underline{88.7} & \underline{66.5} &  \underline{74.2} & \textbf{84.5} & 45.1 & 4.49 \\
        PoseDiffusion\textsuperscript{\dag} & 79.9 & 86.7 & \underline{66.5} & \textbf{79.8} & 80.5 & \textbf{47.5} & 8.83 \\ 
        PoseDiffusion w/o GGS & 72.8 & 82.8 & 56.0 & 67.8 & 75.2 & 43.3 & 0.39 \\
        QuacamFM-\textup{\texttt{lerp}} & 77.2 & 75.8 & 51.3 &  60.4 & 67.8 & 41.3 & 0.46\\
        QuacamFM  & \textbf{86.7} & \textbf{90.5} & \textbf{67.3} &  \underline{74.2} & \underline{82.0} & \underline{45.1} & 0.46 \\
        \bottomrule
    \end{tabularx}
\end{table*}
\paragraph{Metrics.}
To evaluate the performance, we use Relative Rotation Accuracy (RRA) and Relative Translation Accuracy (RTA).
The RRA is used to compare the relative rotation $R_iR_j^T$ from $i$-th camera to $j$-the camera to the ground truth $R_i^*{R_j^*}^T$. 
The RTA defines the angle between the predicted translation and the ground-truth $\btau_{ij}/\btau_{ij}^*$ pointing from camera $i$ to camera $j$ as RTA$(\btau_{ij},{\btau_{ij}}^*)=\texttt{arccos}({\btau_{ij}}^T{\btau_{ij}}^*/\lVert\btau_{ij}\rVert \lVert{\btau_{ij}}^* \rVert)$.
Both metrics are invariant to the choice of global coordinate frame.
Following~\cite{posediffusion,imagematching}, we also report the Area Under the Curve (AUC) metric, which summarizes pose accuracy over a range of angular thresholds for both RRA and RTA.
AUC defines the accuracy at a threshold $\delta$ as $\texttt{min}(\textup{RTA}@\delta, \textup{RRA}@\delta)$.

\paragraph{Baselines.}
We compare against COLMAP+SPSG~\citep{colmap_sfm}, using robust SuperPoint~\citep{superpoint} keypoint detector and SuperGlue~\citep{superglue} matcher.
PixSfM~\citep{pixelsfm} improves reconstruction accuracy through direct alignment of deep features.
For regression-based estimation, we include PoseReg, following the regression baseline introduced in~\cite{posediffusion}.
We also evaluate against RelPose~\citep{relpose}, which models a distribution over camera rotations, and RelPose++~\citep{relpose++}, which extends RelPose to jointly estimate camera translations.
Importantly, we compare with PoseDiffusion~\citep{posediffusion}, a state-of-the-art probabilistic method for sparse-view camera pose estimation that incorporates geometric guidance on correspondences during the diffusion sampling process.
For a fair comparison, we also report PoseDiffusion without geometric guidance (PoseDiffusion w/o GGS), which relies solely on the learned diffusion model and does not require explicit correspondence. 
Methods that require post optimization or access to correspondences are marked with \textsuperscript{\dag}.

We train our method only on CO3Dv2 and evaluate it on the test set, unseen (held-out) CO3Dv2 categories, and RealEstate10K. Results are reported in~\cref{subsec:result_pose}.
To further highlight the contribution of our quaternion-constrained flow formulation, we compare QuacamFM against a vanilla flow-matching baseline, QuacamFM-\texttt{lerp}, in~\cref{subsec:slerp_vs_lerp}.
\subsection{Pose Estimation}
\label{subsec:result_pose}
\paragraph{Result on CO3Dv2.} 

We compare QuacamFM with baseline methods on CO3Dv2 using varying numbers of input images.
\cref{fig:quantitative_co3d} presents quantitative results for 3 to 20 input views, while~\cref{tab:result_co3d} reports camera pose accuracy with 10 input images, and the inference time is measured on a \texttt{NVIDIA-A100} GPU.
Our method substantially outperforms the SfM-based approaches, COLMAP+SPSG and PixSfM, as well as the direct regression baseline, PoseReg.
It further achieves the best performance across all metrics at the threshold of $@30$ in both sparse- and dense-view settings, while remaining competitive at the stricter threshold $@15$.
PoseDiffusion achieves higher translation accuracy on RTA@15 due to its geometric guidance (GGS) over correspondences, but costs additional computation from iterative optimization over the last 10 denoising steps as in the standard setting~\citep{posediffusion}, \eg significant increase from $0.39\mathrm{s}$ to $8.83\mathrm{s}$.
%
RelPose++ achieves relatively higher rotation accuracy at RRA@15, but also requires an additional optimization stage to enforce mutual geometric consistency among the predicted camera rotations.
We use a common choice of 50 optimization iterations~\citep{camera_as_rays}, taking approximately~$4.49\mathrm{s}$. 
A fairer comparison is PoseDiffusion w/o GGS, which has a similar model capacity and does not have access to correspondences, consuming an approximate similar runtime of $0.4\mathrm{s}$ with our method. 
Our method significantly outperforms this baseline, as shown in~\cref{fig:quantitative_co3d}.
Furthermore, \cref{tab:result_co3d} clearly shows that QuacamFM achieves substantially better camera accuracy than its variant, QuacamFM-\texttt{lerp}.
This demonstrates that intuitively applying a standard flow-matching pipeline to camera pose estimation is not effective, while  explicitly imposing the quaternion constraint leads to significantly improved performance.
Further detailed analysis of \texttt{slerp} versus \texttt{lerp} is provided in~\cref{subsec:slerp_vs_lerp}.

%
\paragraph{Generalization to unseen CO3Dv2 and RealEstate10K.} 
\begin{table}[t]
\centering
%
\begin{minipage}[t]{0.48\linewidth}
\centering
\footnotesize
\setlength{\tabcolsep}{1pt}
\caption{Results on the \textbf{held-out} CO3Dv2 categories with 10 input frames.}
\label{tab:unseenCO3D}
\begin{tabular}{lccc}
\toprule
Method & RTA@15 & RRA@15 & AUC@30 \\
\midrule
COLMAP + SPSG & 43.6 & 35.8 & 30.3 \\
PixSfM & 38.3 & 38.9 & 34.2 \\
PoseReg & 42.2 & 45.5 & 36.4 \\
PoseDiffusion\textsuperscript{\dag} & \textbf{55.8} & \underline{62.8} & \underline{50.8} \\
PoseDiffusion {\scriptsize w/o GGS} & 48.8 & 56.1  & 40.1 \\
QuacamFM-\textup{\texttt{lerp}} & 54.6 & 60.3 & 45.5\\
QuacamFM & \textbf{55.8} & \textbf{64.9} & \textbf{51.1} \\
\bottomrule
\end{tabular}
\end{minipage}
\hfill
\begin{minipage}[t]{0.48\linewidth}
\centering
\footnotesize
\setlength{\tabcolsep}{1pt}
\caption{Results on the RealEstate10K with 10 input frames.}
\label{tab:RE10k}
\begin{tabular}{lccc}
\toprule
Method & RTA@15 & RRA@15 & AUC@30 \\
\midrule
COLMAP + SPSG & 60.2 & 72.8 & 45.2 \\
PixSfM &  \underline{64.2} & \textbf{78.3} & \textbf{49.4} \\
PoseReg & 35.4 & 55.6 & 35.2 \\
PoseDiffusion\textsuperscript{\dag} & \textbf{67.2} & \underline{76.8} & \underline{48.0} \\
PoseDiffusion {\scriptsize w/o GGS} & 38.7 & 64.3 & 28.7 \\
QuacamFM-\textup{\texttt{lerp}} & 50.1& 70.6 & 40.3 \\
QuacamFM     & 50.6 & \textbf{78.3} & 42.7 \\
\bottomrule
\end{tabular}
\end{minipage}
\end{table}
We assess the generalization ability of our method on held out CO3Dv2 object categories and the RealEstate10K dataset.
RealEstate10K is scene-centric and follows a data distribution that is unseen during training.
It contains free-form camera motion, in contrast to the object-centric orbital camera motion in CO3Dv2. 
Following the evaluation protocol in~\cite{posediffusion}, we report the results in~\cref{tab:unseenCO3D} and~\ref{tab:RE10k}.
Our method remains on par with PoseDiffusion, which has access to image correspondences, and outperforms the other baselines.
Qualitative examples on RealEstate10K dataset can be seen in the Appendix. 
%
\subsection{Importance of Quaternion-constrained flows}
\label{subsec:slerp_vs_lerp}
\begin{figure*}[t]
    \centering
    \includegraphics[width=0.99\textwidth]{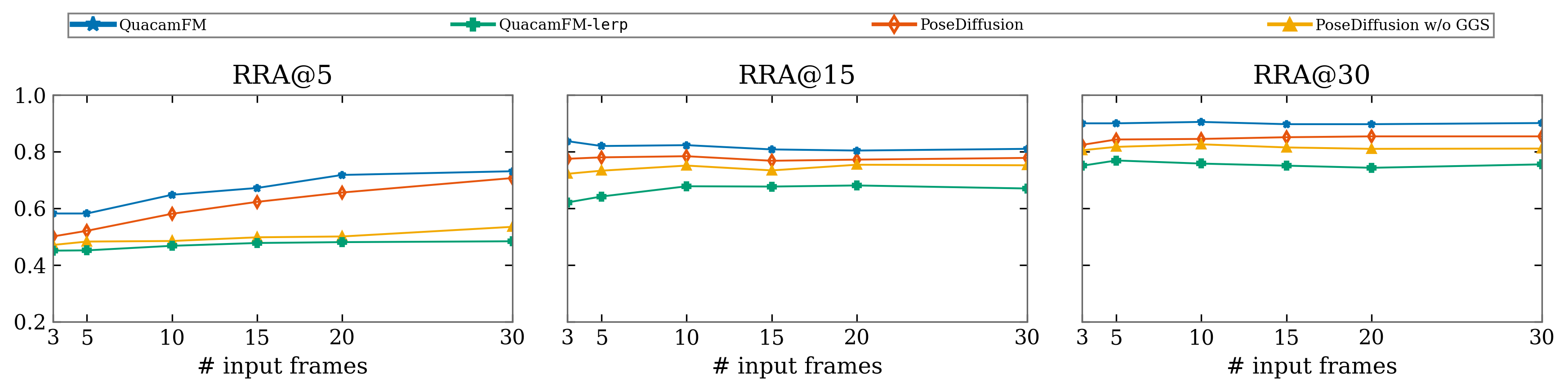}
    \caption{\textbf{Optimal transport ablation}. Comparison of rotation accuracy between quaternion-constrained flows (QuacamFM), vanilla normalizing flows (QuacamFM-\texttt{lerp}), and diffusion-based methods (PoseDiffusion and PoseDiffusion w/o GGS). Experiments conducted on CO3Dv2 test set.}
    \label{fig:slerp_lerp}
\end{figure*}
%
To validate the importance of quaternion-constrained flow, we provide the variant QuacamFM-\texttt{lerp} using the same architecture and training hyper-parameters, but ablating the rotation optimal transport path. 
We mainly evaluate the performance on rotation accuracy, using stricter thresholds (@5, @15, @30) with a wider range of input views to assess performance consistency.
We also compare against other probabilistic methods, \ie PoseDiffusion and PoseDiffusion w/o GGS, in~\cref{fig:slerp_lerp}.

\cref{fig:slerp_lerp} demonstrates that naively treating quaternions as unconstrained 4D vectors (QuacamFM-\texttt{lerp}) in the flow-matching framework does not yield any improvements over the straightforward diffusion-based approach, \ie PoseDiffusion w/o GGS. 
It remains notably behind PoseDiffusion, which has access to image pairwise correspondences during denoising process.
QuacamFM, imposing the unit-constraints and using QDE, shows significant rotation accuracy improvement consistently under multiple thresholds and input image ranges, without access to correspondences.
The camera evolvement animation during integration of QuacamFM, QuacamFM-\texttt{lerp} and during denoising of PoseDiffusion is illustrated in Appendix materials.
\subsection{Demonstration on In-the-wild inputs}
\begin{figure}[h]
    \centering
    \includegraphics[width=0.95\textwidth]{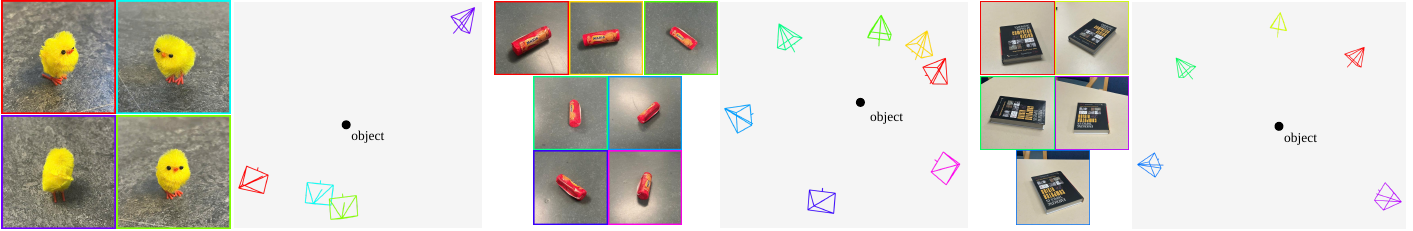}
    \caption{\textbf{Camera pose estimations on in-the-wild examples}. QuacamFM generalizes well to images outside of the distribution of CO3Dv2 training categories, and can recover plausible camera poses even on self-captures that contain varying distances from the camera and in-plane rotations.}
    \label{fig:in-the-wild}
\end{figure}

Finally, we further demonstrate generalization beyond the training data distribution by presenting results on in-the-wild self-captured sequences.
We show results with varying numbers of input images and demonstrate that our method can recover plausible camera poses that are visually consistent with the input views, as shown in~\cref{fig:in-the-wild}.

\section{Conclusion}
This paper presents QuacamFM, a quaternion-constrained flow-matching framework for sparse multi-view camera pose estimation.
QuacamFM uses $\texttt{slerp}$ to design the optimal transport path, thus preserving valid camera rotations throughout the flow trajectory.
By using integration of the quaternion differential equation, it enables smooth and geometrically valid rotational flows.
Our experiments show QuacamFM outperforms diffusion-based approaches and traditional SfM methods, such as COLMAP.
We further demonstrate that explicitly enforcing the quaternion constraint improves sparse-view camera pose estimation compared with naively applying standard flow matching to unconstrained 4D quaternion vectors.
QuacamFM also generalizes well across different datasets and scene distributions.

\noindent{\textit{Limitations and future work.}}
While our method explicitly enforces rotation validity during learning, it have not incorporated epipolar geometric constraints due to the expensive correspondence extraction and optimization processes.
Future work could integrate epipolar geometry directly into camera pose inference for more accurate estimations, while maintaining low computational overhead.
%

\clearpage
\bibliography{iclr2027_conference}

@article{rome,
  title={Building rome in a day},
  author={Agarwal, Sameer and Furukawa, Yasutaka and Snavely, Noah and Simon, Ian and Curless, Brian and Seitz, Steven M and Szeliski, Richard},
  journal={Communications of the ACM},
  volume={54},
  number={10},
  pages={105--112},
  year={2011},
  publisher={ACM New York, NY, USA}
}

@inproceedings{colmap_sfm,
  title={Structure-from-motion revisited},
  author={Schonberger, Johannes L and Frahm, Jan-Michael},
  booktitle={IEEE/CVF Conference on Computer Vision and Pattern Recognition},
  pages={4104--4113},
  year={2016}
}

@article{orbslam2,
  title={ORB-SLAM2: An Open-Source SLAM System for Monocular, Stereo, and RGB-D Cameras},
  author={Mur-Artal, Raul and Tardos, Juan D.},
  journal={IEEE Transactions on Robotics},
  volume={33},
  number={5},
  pages={1255--1262},
  year={2017},
  publisher={IEEE},
  doi={10.1109/TRO.2017.2705103}
}

@inproceedings{superglue,
  title={Superglue: Learning feature matching with graph neural networks},
  author={Sarlin, Paul-Edouard and DeTone, Daniel and Malisiewicz, Tomasz and Rabinovich, Andrew},
  booktitle={IEEE/CVF Conference on Computer Vision and Pattern Recognition},
  pages={4937--4946},
  year={2020},
}

@inproceedings{roma,
  title={Roma: Robust dense feature matching},
  author={Edstedt, Johan and Sun, Qiyu and B{\"o}kman, Georg and Wadenb{\"a}ck, M{\aa}rten and Felsberg, Michael},
  booktitle={IEEE/CVF Conference on Computer Vision and Pattern Recognition},
  pages={19790--19800},
  year={2024},
}

@inproceedings{posediffusion,
  title={Posediffusion: Solving pose estimation via diffusion-aided bundle adjustment},
  author={Wang, Jianyuan and Rupprecht, Christian and Novotny, David},
  booktitle={IEEE/CVF International Conference on Computer Vision},
  pages={9739--9749},
  year={2023}
}

@inproceedings{sparsepose,
  title={Sparsepose: Sparse-view camera pose regression and refinement},
  author={Sinha, Samarth and Zhang, Jason Y and Tagliasacchi, Andrea and Gilitschenski, Igor and Lindell, David B},
  booktitle={IEEE/CVF Conference on Computer Vision and Pattern Recognition},
  pages={21349--21359},
  year={2023}
}

@inproceedings{dust3r,
  title={Dust3r: Geometric 3d vision made easy},
  author={Wang, Shuzhe and Leroy, Vincent and Cabon, Yohann and Chidlovskii, Boris and Revaud, Jerome},
  booktitle={IEEE/CVF Conference on Computer Vision and Pattern Recognition},
  pages={20697--20709},
  year={2024},
}

@inproceedings{transformerpose,
  title={Learning multi-scene absolute pose regression with transformers},
  author={Shavit, Yoli and Ferens, Ron and Keller, Yosi},
  booktitle={IEEE/CVF International Conference on Computer Vision},
  pages={2713--2722},
  year={2021},
  organization={IEEE}
}

@inproceedings{posenet,
  title={Posenet: A convolutional network for real-time 6-dof camera relocalization},
  author={Kendall, Alex and Grimes, Matthew and Cipolla, Roberto},
  booktitle={IEEE International Conference on Computer Vision},
  pages={2938--2946},
  year={2015}
}

@inproceedings{relpose,
  title={Relpose: Predicting probabilistic relative rotation for single objects in the wild},
  author={Zhang, Jason Y and Ramanan, Deva and Tulsiani, Shubham},
  booktitle={IEEE European Conference on Computer Vision},
  pages={592--611},
  year={2022},
}

@misc{flowmatching,
      title={Flow Matching for Generative Modeling}, 
      author={Yaron Lipman and Ricky T. Q. Chen and Heli Ben-Hamu and Maximilian Nickel and Matt Le},
      year={2023},
      eprint={2210.02747},
      archivePrefix={arXiv},
      primaryClass={cs.LG},
      url={https://arxiv.org/abs/2210.02747}, 
}

@article{tong2023improving,
  title={Improving and generalizing flow-based generative models with minibatch optimal transport},
  author={Tong, Alexander and Fatras, Kilian and Malkin, Nikolay and Huguet, Guillaume and Zhang, Yanlei and Rector-Brooks, Jarrid and Wolf, Guy and Bengio, Yoshua},
  journal={arXiv preprint arXiv:2302.00482},
  year={2023}
}

@inproceedings{zhang2025towards,
  title={Towards hierarchical rectified flow},
  author={Zhang, Yichi and Yan, Yici and Schwing, Alex and Zhao, Zhizhen},
  booktitle={International Conference on Learning Representations},
  pages={41537--41558},
  year={2025}
}

@inproceedings{co3d,
  title={Common objects in 3d: Large-scale learning and evaluation of real-life 3d category reconstruction},
  author={Reizenstein, Jeremy and Shapovalov, Roman and Henzler, Philipp and Sbordone, Luca and Labatut, Patrick and Novotny, David},
  booktitle={IEEE/CVF International Conference on Computer Vision},
  pages={10881--10891},
  year={2021},
  organization={IEEE}
}

@inproceedings{coco,
  title={Microsoft coco: Common objects in context},
  author={Lin, Tsung-Yi and Maire, Michael and Belongie, Serge and Hays, James and Perona, Pietro and Ramanan, Deva and Doll{\'a}r, Piotr and Zitnick, C Lawrence},
  booktitle={European Conference on Computer Vision},
  pages={740--755},
  year={2014},
  organization={Springer}
}

@article{re10k,
  title={Stereo magnification: Learning view synthesis using multiplane images},
  author={Zhou, Tinghui and Tucker, Richard and Flynn, John and Fyffe, Graham and Snavely, Noah},
  journal={arXiv preprint arXiv:1805.09817},
  year={2018}
}

@article{orb-slam,
  title={ORB-SLAM: A versatile and accurate monocular SLAM system},
  author={Mur-Artal, Raul and Montiel, Jose Maria Martinez and Tardos, Juan D},
  journal={IEEE transactions on robotics},
  volume={31},
  number={5},
  pages={1147--1163},
  year={2015},
}

@inproceedings{wiles2020synsin,
  title={Synsin: End-to-end view synthesis from a single image},
  author={Wiles, Olivia and Gkioxari, Georgia and Szeliski, Richard and Johnson, Justin},
  booktitle={IEEE/CVF Conference on Computer Vision and Pattern Recognition},
  pages={7465--7475},
  year={2020},
}

@inproceedings{superpoint,
  title={Superpoint: Self-supervised interest point detection and description},
  author={DeTone, Daniel and Malisiewicz, Tomasz and Rabinovich, Andrew},
  booktitle={IEEE/CVF Conference on Computer Vision and Pattern Recognition Workshops},
  pages={337--33712},
  year={2018},
}

@inproceedings{pixelsfm,
  title={Pixel-perfect structure-from-motion with featuremetric refinement},
  author={Lindenberger, Philipp and Sarlin, Paul-Edouard and Larsson, Viktor and Pollefeys, Marc},
  booktitle={IEEE/CVF International Conference on Computer Vision},
  pages={5967--5977},
  year={2021},
}

@inproceedings{relpose++,
  title={Relpose++: Recovering 6d poses from sparse-view observations},
  author={Lin, Amy and Zhang, Jason Y and Ramanan, Deva and Tulsiani, Shubham},
  booktitle={International Conference on 3D Vision},
  pages={106--115},
  year={2024},
  organization={IEEE}
}

@article{imagematching,
  title={Image matching across wide baselines: From paper to practice},
  author={Jin, Yuhe and Mishkin, Dmytro and Mishchuk, Anastasiia and Matas, Jiri and Fua, Pascal and Yi, Kwang Moo and Trulls, Eduard},
  journal={International Journal of Computer Vision},
  volume={129},
  number={2},
  pages={517--547},
  year={2021},
  publisher={Springer}
}

@inproceedings{dino,
  title={Emerging properties in self-supervised vision transformers},
  author={Caron, Mathilde and Touvron, Hugo and Misra, Ishan and J{\'e}gou, Herv{\'e} and Mairal, Julien and Bojanowski, Piotr and Joulin, Armand},
  booktitle={IEEE/CVF International Conference on Computer Vision},
  pages={9630--9640},
  year={2021},
}

@article{transformer,
  title={Attention is all you need},
  author={Vaswani, Ashish and Shazeer, Noam and Parmar, Niki and Uszkoreit, Jakob and Jones, Llion and Gomez, Aidan N and Kaiser, {\L}ukasz and Polosukhin, Illia},
  journal={Advances in neural information processing systems},
  volume={30},
  year={2017}
}

@article{sola2017quaternion,
  title={Quaternion kinematics for the error-state Kalman filter},
  author={Sola, Joan},
  journal={arXiv preprint arXiv:1711.02508},
  year={2017}
}

@inproceedings{dedode,
  title={Dedode: Detect, don’t describe—describe, don’t detect for local feature matching},
  author={Edstedt, Johan and B{\"o}kman, Georg and Wadenb{\"a}ck, M{\aa}rten and Felsberg, Michael},
  booktitle={International Conference on 3D Vision},
  pages={148--157},
  year={2024},
  organization={IEEE}
}

@inproceedings{loma,
  title={Loma: Local feature matching revisited},
  author={Nordstr{\"o}m, David and Edstedt, Johan and B{\"o}kman, Georg and Astermark, Jonathan and Heyden, Anders and Larsson, Viktor and Wadenb{\"a}ck, M{\aa}rten and Felsberg, Michael and Kahl, Fredrik},
  booktitle={European Conference on Computer Vision},
  pages={637--655},
  year={2026},
  organization={Springer}
}

@inproceedings{cui2015global,
  title={Global structure-from-motion by similarity averaging},
  author={Cui, Zhaopeng and Tan, Ping},
  booktitle={IEEE International Conference on Computer Vision},
  pages={864--872},
  year={2015}
}

@inproceedings{moulon2013global,
  title={Global fusion of relative motions for robust, accurate and scalable structure from motion},
  author={Moulon, Pierre and Monasse, Pascal and Marlet, Renaud},
  booktitle={IEEE International Conference on Computer Vision},
  pages={3248--3255},
  year={2013}
}

@inproceedings{wilson2014robust,
  title={Robust global translations with 1dsfm},
  author={Wilson, Kyle and Snavely, Noah},
  booktitle={European conference on computer vision},
  pages={61--75},
  year={2014},
  organization={Springer}
}

@article{snavely2006photo,
  title={Photo tourism: exploring photo collections in 3D},
  author={Snavely, Noah and Seitz, Steven M and Szeliski, Richard},
  journal = {ACM Transactions on Graphics},
  volume = {25},
  number = {3},
  pages={835--846},
  year={2006}
}

@inproceedings{wu2013towards,
  title={Towards linear-time incremental structure from motion},
  author={Wu, Changchang},
  booktitle={International Conference on 3D Vision},
  pages={127--134},
  year={2013},
  organization={IEEE}
}

@inproceedings{hloc,
  title={From coarse to fine: Robust hierarchical localization at large scale},
  author={Sarlin, Paul-Edouard and Cadena, Cesar and Siegwart, Roland and Dymczyk, Marcin},
  booktitle={IEEE/CVF Conference on Computer Vision and Pattern Recognition},
  pages={12708--12717},
  year={2019},
}

@inproceedings{vggt,
  title={Vggt: Visual geometry grounded transformer},
  author={Wang, Jianyuan and Chen, Minghao and Karaev, Nikita and Vedaldi, Andrea and Rupprecht, Christian and Novotny, David},
  booktitle={IEEE/CVF Conference on Computer Vision and Pattern Recognition},
  pages={5294--5306},
  year={2025},
}

@inproceedings{vggsfm,
  title={Vggsfm: Visual geometry grounded deep structure from motion},
  author={Wang, Jianyuan and Karaev, Nikita and Rupprecht, Christian and Novotny, David},
  booktitle={IEEE/CVF Conference on Computer Vision and Pattern Recognition},
  pages={21686--21697},
  year={2024},
}

@inproceedings{light3r-sfm,
  title={Light3r-sfm: Towards feed-forward structure-from-motion},
  author={Elflein, Sven and Zhou, Qunjie and Leal-Taix{\'e}, Laura},
  booktitle={IEEE/CVF Conference on Computer Vision and Pattern Recognition},
  pages={16774--16784},
  year={2025},
}

@inproceedings{zhang2025flare,
  title={Flare: Feed-forward geometry, appearance and camera estimation from uncalibrated sparse views},
  author={Zhang, Shangzhan and Wang, Jianyuan and Xu, Yinghao and Xue, Nan and Rupprecht, Christian and Zhou, Xiaowei and Shen, Yujun and Wetzstein, Gordon},
  booktitle={IEEE/CVF Conference on Computer Vision and Pattern Recognition},
  pages={21936--21947},
  year={2025},
}

@inproceedings{
    duisterhof2025mastrsfm,
    title={{MAS}t3R-SfM: a Fully-Integrated Solution for Unconstrained Structure-from-Motion},
    author={Bardienus Pieter Duisterhof and Lojze Zust and Philippe Weinzaepfel and Vincent Leroy and Yohann Cabon and Jerome Revaud},
    booktitle={International Conference on 3D Vision},
    year={2025},
    url={https://openreview.net/forum?id=5uw1GRBFoT}
}

@misc{mast3r_eccv24,
      title={Grounding Image Matching in 3D with MASt3R}, 
      author={Vincent Leroy and Yohann Cabon and Jerome Revaud},
      booktitle = {European Conference on Computer Vision},
      year = {2024},
      organization={Springer}
}

@inproceedings{camera_as_rays,
  title={Cameras as rays: Pose estimation via ray diffusion},
  author={Zhang, Jason and Lin, Amy and Kumar, Moneish and Yang, Tzu-Hsuan and Ramanan, Deva and Tulsiani, Shubham},
  booktitle={International Conference on Learning Representations},
  volume={2024},
  pages={23345--23366},
  year={2024}
}

@inproceedings{zhao2025diffusionsfm,
  title={{DiffusionSfM}: Predicting structure and motion via ray origin and endpoint diffusion},
  author={Zhao, Qitao and Lin, Amy and Tan, Jeff and Zhang, Jason Y and Ramanan, Deva and Tulsiani, Shubham},
  booktitle={IEEE/CVF Conference on Computer Vision and Pattern Recognition},
  pages={6317--6326},
  year={2025},
}

@article{ho2020denoising,
  title={Denoising diffusion probabilistic models},
  author={Ho, Jonathan and Jain, Ajay and Abbeel, Pieter},
  journal={Advances in neural information processing systems},
  volume={33},
  pages={6840--6851},
  year={2020}
}

@book{hartley2003multiple,
  title={Multiple view geometry in computer vision},
  author={Hartley, Richard and Zisserman, Andrew and others},
  year={2003},
  publisher={Cambridge university press Cambridge}
}

@article{ozyecsil2017survey,
  title={A survey of structure from motion*.},
  author={{\"O}zye{\c{s}}il, Onur and Voroninski, Vladislav and Basri, Ronen and Singer, Amit},
  journal={Acta Numerica},
  volume={26},
  pages={305--364},
  year={2017},
  publisher={Cambridge University Press}
}

@inproceedings{brahmbhatt2018geometry,
  title={Geometry-aware learning of maps for camera localization},
  author={Brahmbhatt, Samarth and Gu, Jinwei and Kim, Kihwan and Hays, James and Kautz, Jan},
  booktitle={2018 IEEE/CVF conference on Computer Vision and Pattern Recognition},
  pages={2616--2625},
}

@book{dunn2011primer,
    author = {Dunn, Fletcher and Parbery, Ian},
    title = {3D Math Primer for Graphics and Game Development},
    publisher = {Taylor \& Francis Inc.},
    year = {2011},
}

@inproceedings{le2026quamo,
    title     = {QuaMo: Quaternion Motions for Vision-based 3D Human Kinematics Capture},
    author    = {Le, Cuong and Melnyk, Pavlo and Waldmann, Urs and Wadenbäck, Mårten and Wandt, Bastian},
    booktitle = {International Conference on Learning Representations},
    year      = {2026},
    url       = {https://openreview.net/forum?id=em0jPLYjaS},
}

@book{kuipers1999quaternions,
    author    = {Kuipers, Jack B},
    title     = {Quaternions and rotation sequences: a primer with applications to orbits, aerospace, and virtual reality},
    publisher = {Princeton university press},
    year      = {1999}}

@article{golabek2022aircraft,
    author    = {Gołąbek, Michał and Welcer, Michał and Szczepański, Cezary and Krawczyk, Mariusz and Zajdel, Albert and Borodacz, Krystian},
    title     = {Quaternion Attitude Control System of Highly Maneuverable Aircraft},
    journal   = {Electronics},
    volume    = {11},
    year      = {2022},
    number    = {22},
    article-number = {3775},
    ISSN      = {2079-9292},
    DOI       = {10.3390/electronics11223775}}

@inproceedings{he2016deep,
  title={Deep residual learning for image recognition},
  author={He, Kaiming and Zhang, Xiangyu and Ren, Shaoqing and Sun, Jian},
  booktitle={IEEE/CVF Conference on Computer Vision and Pattern Recognition},
  pages={770--778},
  year={2016}
}
\bibliographystyle{iclr2027_conference}

\clearpage
\appendix
\section{Appendix}
We provide supplementary videos visualizing the evolution of the estimated camera poses throughout the inference process, from their initial states toward the final predictions.
These videos show QuacamFM, QuacamFM-\texttt{lerp}, and PoseDiffusion.
These visualizations can be accessed through the provided anonymous local webpage, \texttt{quacamfm\_appendix.html}.

\paragraph{Qualitative result on RealEstate10K.}
\begin{figure*}[t]
    \centering
    \includegraphics[width=0.99\textwidth]{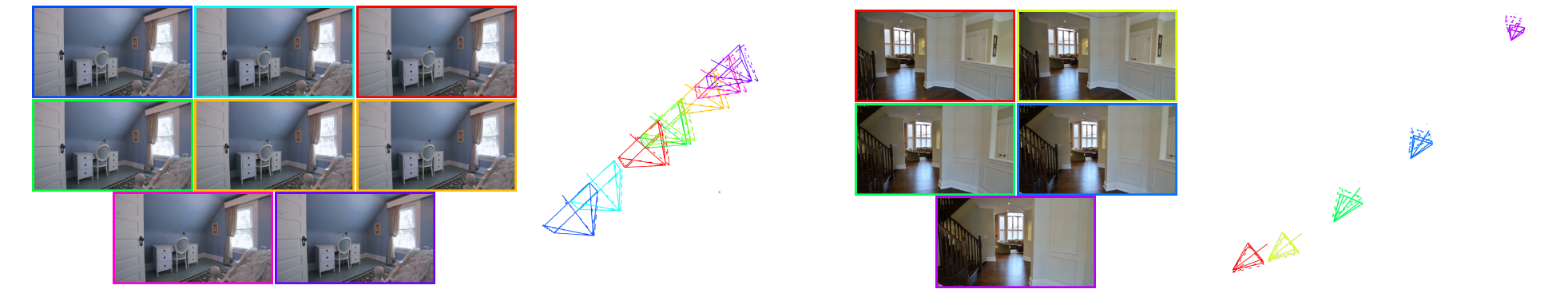}
    \caption{Qualitative results of QuacamFM on RealEstate10K.}
    \label{fig:qualitative_re10k}
\end{figure*}

We provide qualitative examples of QuacamFM on the RealEstate10K dataset in~\cref{fig:qualitative_re10k}. 
The examples cover diverse camera motions, including forward motion combined with rotation and side-to-side motion.

\noindent \textbf{Ablation study.}
We investigate the effect of different visual feature extractors, as in~\cref{tab:backbones}, by evaluating three backbone configurations: 
ResNet50, trained by supervised image classification~\citep{he2016deep};
ResNet50-DINO~\citep{dino}, trained in a self-supervised manner;
and the ViT-DINO~\citep{dino} model.
We ablate the number of integration steps used to solve the flow-matching ODE during inference as in~\cref{tab:timesteps}.
All ablation experiments are conducted on CO3Dv2.
\begin{table}[h]
\centering
\begin{minipage}[t]{0.48\linewidth}
\centering
\scriptsize
\setlength{\tabcolsep}{4pt}
\caption{QuacamFM performance with different visual extractor backbones, evaluated with 10-image input.}
\label{tab:backbones}
\begin{tabular}{lccc}
\toprule
Backbones & ResNet50 & ResNet50-DINO & ViT-DINO \\
\midrule
RTA@15 & 44.1 & 43.6 & \textbf{45.1}  \\
RRA@15 & 77.9 & 78.4 & \textbf{82.0}  \\
AUC@30 & 62.7 & 64.2 & \textbf{67.3}  \\
\bottomrule
\end{tabular}
\end{minipage}
\hfill
%
\begin{minipage}[t]{0.48\linewidth}
\centering
\scriptsize
\setlength{\tabcolsep}{3pt}
\caption{QuacamFM performance with different number of timesteps, using the ViT-DINO feature backbone.}
\label{tab:timesteps}
\begin{tabular}{lcccc}
\toprule
Metrics & 20 steps & 30 steps & 50 steps & 80 steps \\
\midrule
RTA@15 & 43.9 & \textbf{45.1} & 43.6 & 44.2 \\
RRA@15 & 79.5 & \textbf{82.0} & 81.0 & 79.8 \\
AUC@30 & 64.2 & \textbf{67.3} & 65.8 & 66.4 \\
\bottomrule
\end{tabular}
\end{minipage}
\end{table}

\paragraph{Inference time.}
In~\cref{tab:time_consume}, we report the inference time of QuacamFM, PoseDiffusion w/o GGS, PoseDiffusion, and RelPose++ with a different number of input images. It is measured on a single NVIDIA A100-SXM4-40GB GPU. 
This shows how the computation of PoseDiffusion and RelPose++ grows when they incorporate epipolar constraint-based optimization and camera rotation consistency optimization.
We configure PoseDiffusion to apply the standard GGS during the last 10 denoising steps, and RelPose++ uses its commonly adopted setting of 50 optimization iterations.
\begin{table}[h]
\centering
\setlength{\tabcolsep}{4pt}
\caption{Inference time of methods with multi-image inputs. The inference time is measured in second $\mathrm{s}$.}
\label{tab:time_consume}
\begin{tabular}{lccc}
\toprule
Method & 5 images & 10 images & 20 images \\
\midrule
PoseDiffusion & 6.344 & 8.834 & 20.61 \\
PoseDiffusion w/o GGS & 0.342 & 0.393 & 0.825  \\
RelPose++ & 2.854 & 4.491 & 11.32 \\
QuacamFM & 0.352 & 0.465 & 0.735  \\
\bottomrule
\end{tabular}
\end{table}

\end{document}